# A Resolution Calculus for Dynamic Semantics

Christof Monz and Maarten de Rijke

ILLC, University of Amsterdam, Plantage Muidergracht 24, 1018 TV Amsterdam, The
Netherlands. E-mail: {christof, mdr}@wins.uva.nl

**Abstract.** This paper applies resolution theorem proving to natural language semantics. The aim is to circumvent the computational complexity triggered by natural language ambiguities like pronoun binding, by interleaving pronoun binding with resolution deduction. To this end, disambiguation is only applied to expressions that actually occur during derivations. Given a set of premises and a conclusion, our resolution method only delivers pronoun bindings that are needed to derive the conclusion.

## 1 Introduction

Natural language processing (NLP), has a long tradition in Artificial Intelligence, but it still remains to be one of the hardest problems in the area. Research areas such as semantic representation and theorem proving with natural language have to deal with a problem that is characteristic of natural languages, namely ambiguity. There are several kinds of ambiguity, see for instance [RN95] for an overview. In the present paper, we focus on pronoun binding,[1] a certain instance of ambiguity, as exemplified by (1) below.

(1) A man sees a boy. He whistles.

Often, there are lots of possibilities to bind a pronoun and it is not clear which one to choose. The pronoun *he* in the short discourse in (1) can be bound in two ways as given in (2), where co-indexation indicates referential identity.

(2) a.  A man$_i$ sees a boy. He$_i$ whistles.
    b.  A man sees a boy$_i$. He$_i$ whistles.

For some cases heuristics are applicable which prefer certain bindings to others, but at present there is no approach making use of heuristics which is general enough to cover all problems.

Dynamic semantics [Kam81,GS91] allows to give a perspicuous solution to some problems involving pronoun binding. Since we are interested in binding occurrences of pronouns to expressions mentioned earlier in a discourse, we take a slight modification of Dynamic Predicate Logic (DPL) [GS91], where it is not presupposed that pronouns are already co-indexed. Actually, pronoun binding falls into the realm of constructing semantic representations of natural language discourses, and one of the main purposes of constructing these representations is to reason with them. Now, the question arises which form the input of the theorem prover should have. Should a theorem prover work only on totally disambiguated expressions? Total disambiguation results in an explosion of readings, because of the multiplicative behavior of ambiguity. On the other hand, to prove a conclusion $\varphi$ from a set of premises $\Gamma$ it may be enough to use only premises from a small subset $\Delta$ of $\Gamma$, and it may be sufficient, and much more efficient, to disambiguate only $\Delta$ instead of the whole set of premises $\Gamma$. In general, we do not know in advance which subset of premises might be enough to derive a certain conclusion, but during a derivation often certain (safe) strategies may be applied that prevent some premises from being used since they cannot lead to the conclusion, anyway. Common strategies to constrain the search space in resolution deduction are e.g., the *set-of-support strategy* and *ordered resolution*. Our goal is to constrain the set of premises that have to be disambiguated by interleaving deduction and disambiguation. Roughly speaking, premises are only disambiguated if they are used by a deduction rule.

---

[1] Throughout this paper we use the term *binding* to express the referential identification of a pronoun and another referential expression occurring in the discourse. Common terms are also *co-indexation* or *pronoun resolution*. We especially did not use *pronoun resolution* to avoid confusion with resolution as a deduction principle.

The rest of the paper is structured as follows. Section 2 provides some rudimentary background in dynamic semantics and explains what kind of structural information is necessary to restrict pronoun binding. In addition, the basics of resolution deduction are introduced. Section 3 discusses some of the problems of the (standard) resolution method when applied to natural language. The method of labeled unification and resolution is presented to overcome these problems. Section 4 briefly relates our work to some other approaches to pronoun binding. Section 5 provides some conclusions and prospects for further work.

## 2  Background

Before we turn to our method of labeled resolution deduction and its applications to discourse semantics, we briefly present the idea of dynamic semantics. The second subsection shortly explains the classic resolution method for (static) first-order logic.

### 2.1  Dynamic Reasoning

Dynamic reasoning differs from classical reasoning to the extent that sequences of formulas are considered instead of sets of formulas. To model discourse relations like pronoun binding it is important to take the order of sentences into account because two sequences which have the same members, but differ in order, may have a different meaning. (Compare 'A man walks in the park. He whistles.' and 'He whistles. A man walks in the park.')

DPL is a semantic framework which works on sequences of formulas and it allows to represent pronoun binding, where the antecedent of the pronoun and the pronoun itself may occur in different formulas. This is accomplished by assigning the existential quantifier flexible binding. In (3.b) a DPL representation of the short discourse in (3.a) is given.

(3)  a.  A man$_i$ sees a boy. He$_i$ whistles.

b.  $\exists x\,(man(x) \land \exists y\,(boy(y) \land see(x,y))) \land whistle(x)$

The pronoun *he* is represented by the variable $x$ which is the same as the one bound by the existential quantifier, but it occurs outside of its scope. To bind $x$ in $whistle(x)$ it is necessary to give the existential quantifier flexible scope.

One of the advantages of dynamic approaches like DPL is that they allow for a formal definition of possible antecedents for a pronoun. Without giving too many details, we just note that negations function as barriers for flexible binding. Therefore, an existential quantifier occurring in the scope of a negation cannot bind a pronoun that occurs outside of the negation, as shown by (4).

(4)  *John doesn't own a car$_i$. It$_i$ is in front of his house.

The three properties (a) existential quantifiers can bind variables occurring to the right-hand side of their traditional scope, (b) conjunctions preserve the flexible scope, and (c) negations are barriers for dynamic binding, allow us to define the properties of the other logical connectives $\lor$, $\rightarrow$ and $\forall$. $[\![\cdot]\!]$ is a function that assigns to each formula its semantic value.

(5)
$$[\![\varphi \lor \psi]\!] = [\![\neg(\neg\varphi \land \neg\psi)]\!]$$
$$[\![\varphi \rightarrow \psi]\!] = [\![\neg(\varphi \land \neg\psi)]\!]$$
$$[\![\forall x\, \varphi]\!] = [\![\neg\exists x\, \neg\varphi]\!]$$

Given these definitions, we see that disjunction is a barrier both internally and externally, implication is a barrier externally but internally it allows for flexible binding, and universal quantification does not allow for external binding.

We differ in two respects from DPL. First, we do not allow two or more occurrences of $\exists x$ within a single text. The problem is that the second occurrence of $\exists x$ resets the value of $x$, and thereby previous restrictions on $x$ are lost. We assume for simplicity that all bound variables are disjoint. This is not a severe restriction and an algorithm for constructing semantic representations for natural language sentences can easily accomplish this. The second difference with DPL is that we do not assume co-indexation of quantifiers and the pronouns which they bind. In (3) the variable for *he* is already assumed to be $x$ and in DPL the question of pronoun binding is pushed to some kind of preprocessing. But finding the right binding is far from being an easy task and it is very complex from a computational point of view. The pronoun in (3) could also be represented by $y$, indicating that that *he* refers to *a boy*. E.g., a discourse containing twenty indefinites followed by a sentence with two pronouns, has $20 \cdot 20 = 400$ possible bindings, disregarding any linguistic constraints which rule out some of the bindings.

To this end, we postpone pronoun binding and represent pronouns in the semantic representation by free variables. Variables for pronouns are displayed in boldface and are of a different kind than regular variables. Pronoun variables are bound by the ?-operator. It differs from $\exists$ and $\forall$, because it only binds its argument, but does not quantify over it. Actually, it is not necessary to have a special operator for pronouns, and we only introduced it here for the sake of convenience to identify the position where the pronoun is introduced. Our representation of (1), repeated as (6.a) below, is given in (6.b). As mentioned before, co-indexation of pronouns and antecedents is not carried out.

(6) a. A man sees a boy. He whistles.
    b. $\exists x\,(man(x) \land \exists y\,(boy(y) \land see(x,y))) \land \,?\mathbf{u}\,whistle(\mathbf{u})$

The task whether **u** has to be substituted by $x$ or by $y$ is postponed to the deduction component, as motivated in Section 1.

Unlike the existential quantifier, the ?-operator does not have the property of flexible binding. We get the following equivalence:

$$[\![\neg\,?\mathbf{u}\varphi]\!] \;=\; [\![?\mathbf{u}\neg\varphi]\!]$$

To define *accessibility* we can now say that a variable $x$ is accessible from a pronoun **u** if no barrier occurs between the quantifier introducing $x$ and $?\mathbf{u}$. A formal definition of accessibility is given in the next section. The equations in (5) show that $\lor$, $\to$ and $\forall$ introduce barriers because of the way they are defined in terms of negation. This is exemplified by (7) below.

(7) *Every farmer owns a donkey$_i$. It$_i$ is grey.

Dispensing with the presupposition that pronouns and antecedents are already co-indexed re-introduces the concept of ambiguity to our framework. This makes it necessary to give a definition of the semantics of ambiguous formulas. It is common to define their semantics in terms of their possible disambiguations, see [Rey93], and here we follow the same approach. A *total disambiguation* is a mapping $\delta$ from ambiguous dynamic formulas to classical first-order formulas. Disambiguation encompasses two steps. First, we have to find a proper antecedent for a pronoun. To define proper antecedents, we use the notion of accessibility. Second, we have to map unambiguous dynamic formulas to classical formulas. This means that we have to turn flexible quantification into static quantification, and this involves re-bracketing and quantifier movement. [GS91] give an algorithm that computes for each DPL-formula $\varphi$ a formula $\varphi'$ which is in normal binding form, i.e., all pronouns are quantified over in the classical sense, and which is valid in first-order logic iff $\varphi$ is valid in DPL. For instance, the normal binding form of (8.b) is (9).

(8) a. If a farmer$_i$ owns a donkey$_j$, then he$_i$ beats it$_j$.
    b. $\exists x\,(f(x) \land \exists y\,(d(y) \land o(x,y))) \to b(x,y)$
(9) $\forall x \forall y\,(f(x) \land d(y) \land o(x,y) \to b(x,y))$

To define the validity of ambiguous formulas, we say that an ambiguous formula $\varphi$ is *valid*, i.e., for all models $M$ it holds that $M \models_a \varphi$, if there is a disambiguation $\delta$, such that $M \models \delta(\varphi)$, for all models $M$. In words: $\varphi$ is valid iff there exists a disambiguation which is valid in first-order logic.

Unfortunately we do not have enough space to give a more detailed account of dynamic semantics, but we refer the reader to [Kam81,GS91].

## 2.2 The Resolution Method

The *resolution method* [Rob65] has become quite popular in automated theorem proving, because it is very efficient and it is easily augmentable by lots of strategies which restrict the search space, see e.g., [Lov78]. On the other hand, the resolution method has the disadvantage of presupposing that its input has to be in *clause form*, which is a set of clauses, interpreted as a conjunction. A *clause* is a set of literals, interpreted as a disjunction. Probably the most attractive feature of resolution is that it has only one inference rule, the resolution rule:

$$\frac{C \cup \{\neg P_1, \ldots, \neg P_n\} \quad D \cup \{Q_1, \ldots, Q_m\}}{(C \cup D\pi)\sigma} \ (res)$$

where
- $Q_1, \ldots, Q_m$ are atomic
- $\pi$ is a substitution such that $C \cup \{\neg P_1, \ldots, \neg P_n\}$ and $D\pi \cup \{Q_1\pi, \ldots, Q_m\pi\}$ are variable disjoint
- $\sigma$ is the most general unifier of $\{P_1, \ldots, P_n, Q_1\pi, \ldots, Q_m\pi\}$

To prove that $\Gamma \models \varphi$ holds we transform $(\bigwedge \Gamma) \wedge \neg \varphi$ in clause form and try to derive a contradiction (the empty clause) from it by using the resolution rule.

For a comprehensive introduction to resolution see for instance [Lov78].

## 3 Dynamic Resolution

Applying the classical resolution method to a dynamic semantics causes problems. Below we will first discuss some of them and then see how we have to design our dynamic resolution method to overcome these problems.

### 3.1 Adapting the Resolution Method

There are two problems that we have to find a solution for. First, transforming formulas to clause form causes a loss of structural information. Therefore, it is sometimes impossible to distinguish between variables that can serve as antecedents for a pronoun and variables than can not. The second problem concerns the duplication of literals which may occur during clause from transformation and the assumption of the resolution method that clauses are variable disjoint. Although the same pronoun may have two occurrences in different clauses, we do not want them to be bound by different antecedents.

Turning to the first problem, in (10) the pronoun **u** cannot be bound by the existential quantifier, whereas the pronoun **z** can be bound by it.

(10)a. Every farmer who owns a donkey beats it. It suffers.
   b. $\forall x (f(x) \wedge \exists y (d(y) \wedge o(x,y)) \rightarrow ?\mathbf{z} b(x,\mathbf{z}))) \wedge ?\mathbf{u} s(\mathbf{u})$
(11) $\{ \{\neg f(x), \neg d(y), \neg o(x,y), b(x,\mathbf{z})\}, \{s(\mathbf{u})\} \}$

How can we tell which identifications are allowed by looking at the corresponding clause form in (11)? How do we know whether a term is accessible?

We use *labels* to carry the information about accessible variables. Each pronoun variable is annotated with a label that indicates the set of accessible variables. Besides the set of first-order or proper variables (*VAR*), first-order formulas (*FORM*), and pronoun variables (*PVAR*), we are going to introduce the sets of labeled pronoun variables (*LPVAR*) and labeled formulas (*LFORM*). Labeled pronoun variables are of the form $V:\mathbf{u}$, where $V \subseteq \textit{VAR}$ and $\mathbf{u}$ is a pronoun variable. *LFORM* is the set of first-order formulas plus formulas containing labeled pronoun variables. To be able to recognize the antecedents later on, each variable is annotated with its name, $(x^x, y^y, \ldots)$, and during skolemization only the variable is changed, but the label remains unchanged.

To see which variables inside of a formula $\varphi$ can serve as antecedents for pronouns, [GS91] introduce the function AQV which returns the set of *actively quantifying variables* when applied to $\varphi$.

**Definition 1.** Let *FORM* be the set of classical first-order formulas and *VAR* the set of first-order variables. The function $\mathsf{AQV} : \textit{FORM} \to \textit{POW}(\textit{VAR})$ is defined recursively:

$$\begin{aligned}
\mathsf{AQV}(R(x_1 \ldots x_n)) &= \emptyset \\
\mathsf{AQV}(\neg \varphi) &= \emptyset \\
\mathsf{AQV}(\varphi \wedge \psi) &= \mathsf{AQV}(\varphi) \cup \mathsf{AQV}(\psi) \\
\mathsf{AQV}(\varphi \to \psi) &= \emptyset \\
\mathsf{AQV}(\varphi \vee \psi) &= \emptyset \\
\mathsf{AQV}(\forall x \varphi) &= \emptyset \\
\mathsf{AQV}(\exists x \varphi) &= \mathsf{AQV}(\varphi) \cup \{x\} \\
\mathsf{AQV}(?\mathbf{u}\varphi) &= \mathsf{AQV}(\varphi)
\end{aligned}$$

Using the above definition we define the notion of accessible variables.

**Definition 2 (Annotation with Accessible Variables).** To annotate $\mathbf{u}$ in $?\mathbf{u}\psi$, we drop the binding operator $?\mathbf{u}$ and substitute all occurrences of the pronoun variable in $\psi$ by its annotated counterpart. The annotation function $\mathsf{annot} : \textit{VAR} \times \textit{FORM} \to \textit{LFORM}$ is defined recursively, where $V \subseteq \textit{VAR}$:

$$\begin{aligned}
\mathsf{annot}(V, R(x_1 \ldots x_n)) &= R(x_1 \ldots x_n) \\
\mathsf{annot}(V, \neg \varphi) &= \neg \mathsf{annot}(V, \varphi) \\
\mathsf{annot}(V, \varphi \wedge \psi) &= \mathsf{annot}(V, \varphi) \wedge \mathsf{annot}(V \cup \mathsf{AQV}(\varphi), \psi) \\
\mathsf{annot}(V, \varphi \to \psi) &= \mathsf{annot}(V, \varphi) \to \mathsf{annot}(V \cup \mathsf{AQV}(\varphi), \psi) \\
\mathsf{annot}(V, \varphi \vee \psi) &= \mathsf{annot}(V, \varphi) \vee \mathsf{annot}(V, \psi) \\
\mathsf{annot}(V, \forall x \varphi) &= \forall x\, \mathsf{annot}(V \cup \{x\}, \varphi) \\
\mathsf{annot}(V, \exists x \varphi) &= \exists x\, \mathsf{annot}(V \cup \{x\}, \varphi) \\
\mathsf{annot}(V, ?\mathbf{u}\varphi) &= \mathsf{annot}(V, \varphi[\mathbf{u}/V:\mathbf{u}])
\end{aligned}$$

The actual annotation takes place in the last case, where the pronoun is substituted. The other cases thread the actively quantifying variables through the formula. To annotate a whole discourse $\varphi_1 \wedge \cdots \wedge \varphi_n$, the variable parameter of $\mathsf{annot}$ is initialized with $\emptyset$, $\mathsf{annot}(\emptyset, \varphi_1 \wedge \cdots \wedge \varphi_n)$. A term $t^x$ is *accessible from* a pronoun $\mathbf{u}$ iff $x$ is element of the set of the accessible variables of $\mathbf{u}$.

Reconsider the last example, *every farmer who owns a donkey beats it. It suffers*. Applying annotation yields:[2]

$$\begin{aligned}
&\mathsf{annot}(\emptyset, \forall x\, (f(x) \wedge \exists y\, (d(y) \wedge o(x,y)) \to ?\mathbf{z}\, b(x, \mathbf{z})) \wedge ?\mathbf{u}\, s(\mathbf{u})) \\
&= \forall x\, (f(x) \wedge \exists y\, (d(y) \wedge o(x,y)) \to b(x, \{x,y\}:\mathbf{z}))) \wedge s(\emptyset:\mathbf{u})
\end{aligned}$$

---

[2] For simplicity, we neglect the fact that pronouns and their antecedents have to agree in gender, number, etc.

Applying clause form transformation to the annotated formulas yields:

(12) $\{\{\neg f(x), \neg d(y), \neg o(x,y), b(x, \{x,y\} : \mathbf{z})\}, \{s(\emptyset : \mathbf{u})\}\}$

We can also see that (10.a) is not well-formed because there are no accessible pronouns for the second pronoun *it*, i.e., the label of $\mathbf{u}$ is the empty set.

Now we turn to the second problem: how do we make sure that the same pronoun, occurring in different clauses, is bound to the same antecedent? As we said earlier, we do not want to assume pronouns to be bound in a set of premises when we apply resolution. The reason is that pronoun binding is highly ambiguous and often it is not necessary to bind all pronouns in a set of premises to derive a certain conclusion from it. Another issue, which we briefly hinted at in Section 2, is that pronouns should be treated as free variables of a special kind, not to be dealt with in the same manner as universally quantified variables (which also happen to be represented by free variables). This is illustrated by the following example, which shows an invalid entailment.

(13)a. $\exists x \exists y ((A(x) \lor A(y)) \land (?\mathbf{z}A(\mathbf{z}) \rightarrow (B \land C))) \not\models_a B \lor C$
   b. $\{\{A(f^x), A(g^y)\}, \{\neg A(\mathbf{z}), B\}, \{\neg A(\mathbf{z}), C\}, \{\neg B\}, \{\neg C\}\}$

The transformation in (13) causes a duplication of the literal $\neg A(\mathbf{z})$, and we have to make sure that the pronoun is instantiated the same way in both cases.

(14) $\{A(f^x), A(g^y)\}$ $\{\neg A(\mathbf{z}), B\}$ $\{\neg A(\mathbf{z}), C\}$ $\{\neg B\}$ $\{\neg C\}$

$\{A(g^y), B\}$

$\{B, C\}$

$\{C\}$

$\square$

In (14) $\mathbf{z}$ is instantiated with $f^x$ in the first resolution step and then with $g^y$ in the second. The resolution rule as it was stated in the preceding section assumes that clauses to be resolved are variable disjoint. We have to modify the resolution rule such that the same pronoun variable is allowed to occur in both clauses. Additionally, the instantiation of a pronoun variable for constructing the most general unifier in a resolution step is applied globally, i.e., to all clauses.

(15) $\{A(f^x), A(g^y)\}$ $\{\neg A(\mathbf{z}), B\}$ $\{\neg A(\mathbf{z}), C\}$ $\{\neg B\}$ $\{\neg C\}$

$\{A(g^y), B\}$ $\{A(f^x), C\}$

$\{A(g^y)\}$ $\{A(f^x)\}$

Global instantiation correctly prevents us from deriving a contradiction in (15).

## 3.2 Labeled Resolution

Unification is a fundamental technique in the resolution method. Since we are also dealing with labeled variables, we have to think how the unification mechanism has to be adapted. In the course of this subsection, it will turn out that pronoun binding can be reduced to unification.

**Labeled Unification.** We use the unification algorithm of Martelli and Montanari [MM82] as a basis and adapt it in such a way that it can deal with labeled pronoun variables.

What does it mean to unify a set of equations $E = \{s_1 \approx t_1, \ldots, s_n \approx t_n\}$, where $s_i$ or $t_i$ can also be a labeled pronoun variable? We have to distinguish three possible cases: (i) neither $s_i$ nor $t_i$ is a labeled pronoun variable, then labeled unification and normal unification are the same thing, (ii) one of them is a pronoun and the other is not, and (iii) both are pronouns. Case (ii) is the normal pronoun binding, where one tries to identify a pronoun with a proper variable. Case (iii) is not an instance of pronoun binding, but an identification of two pronouns, i.e., whatever is the antecedent of the first pronoun, it is also the antecedent of the other one.

**Definition 3 (Labeled Unifier).** We call a substitution $\sigma$ a *labeled unifier* or *unifier\** of a set of equations $E = \{s_1 \approx t_1, \ldots, s_n \approx t_n\}$ iff

1. $s_1\sigma = t_1\sigma, \ldots, s_n\sigma = t_n\sigma$
2. if $(V:\mathbf{u})\sigma = t^x$, then $x \in V$
3. if $(V:\mathbf{u})\sigma = V':\mathbf{v}$ then $V' \subseteq V$

We use $\approx$ to express equality in our object language, whereas $=$ denotes equality in the meta language.

Condition 1 is the normal condition of unifiability, namely that the terms of an equation have to be identical after substitution. The second condition says that unifiers have to obey accessibility, for instance $\sigma := [\{x,y\}:\mathbf{u}/g^z]$ is not a unifier of $\{\{x,y\}:\mathbf{u} \approx g^z\}$, because $g^z$ is not accessible from $\mathbf{u}$, as $z \notin \{x,y\}$. To ensure that identification of pronouns always restricts the set of accessible antecedents, we need condition 3.

**Definition 4 (Most General Labeled Unifier).** A labeled unifier $\sigma$ of a set of equations $E = \{s_1 \approx t_1, \ldots, s_n \approx t_n\}$ is the *most general labeled unifier* or *mgu\** of $E$ if

1. if $\theta$ is a unifier\* of $E$ then there is substitution $\tau$ such that $\theta = \sigma\tau$
2. if $(V:\mathbf{u})\sigma = V_1:\mathbf{v}$, $(V:\mathbf{u})\theta = V_2:\mathbf{v}$, $V_1, V_2 \subseteq V$, and $V_1, V_2 \neq \emptyset$
   then $V_2 \subseteq V_1$

Again, the first condition is standard in regular unification. Condition 2 says that the most general unifier\* has to restrict the set of accessible antecedents as little as possible when identifying pronouns. To unify $V_1:\mathbf{u}$ and $V_2:\mathbf{v}$ it suffices to take any non-empty subset of the intersection of $V_1$ and $V_2$, but this fact may prohibit some antecedents from being accessible, although they are in fact accessible for both pronouns.

**Definition 5 (The Labeled Unification Algorithm).** First, the unification function unify$^*$ is applied to a pair of atoms, and then it tries to unify the set of corresponding argument pairs. The algorithm terminates successfully if it did not terminate with failure and no further equations are applicable.

1. unify$^*(R(s_1...s_n), R(t_1..t_n))$
   = unify$^*(\{s_1 \approx t_1...s_n \approx t_n\})$
2. unify$^*(\{f(s_1...s_n) \approx f(t_1...t_n)\} \cup E)$
   = unify$^*(\{s_1 \approx t_1...s_n \approx t_n\} \cup E)$
3. unify$^*(\{f(s_1...s_n) \approx g(t_1...t_m)\} \cup E)$, $f \neq g$ or $n \neq m$
   = terminate with failure
4. unify$^*(\{x \approx x\} \cup E$
   = unify$^*(E)$
5. unify$^*(\{t \approx x\} \cup E)$, $t \notin VAR$
   = unify$^*(\{x \approx t\} \cup E)$
6. unify$^*(\{x \approx t\} \cup E)$, $x \neq t$, $t \notin LPVAR$, $x$ in $t$
   = terminate with failure
7. unify$^*(\{x \approx t\} \cup E)$, $x \neq t$, $t \notin LPVAR$, $x$ not in $t$, $x$ in $E$
   = unify$^*(\{x \approx t\} \cup E[x/t])$
8. unify$^*(\{V:\mathbf{u} \approx t^x\} \cup E)$, $x \in V$, $V:\mathbf{u}$ in $E$
   = unify$^*(\{V:\mathbf{u} \approx t^x\} \cup E[V:\mathbf{u}/t^x])$
9. unify$^*(\{V_1:\mathbf{u} \approx V_2:\mathbf{v}\} \cup E)$, $V_1 \cap V_2 \neq \emptyset$, $V_1 \cap V_2 \subset V_2$
   = unify$^*(\{V_1:\mathbf{u} \approx V_1 \cap V_2:\mathbf{v}, V_2:\mathbf{v} \approx V_1 \cap V_2:\mathbf{v}\}$
   $\cup E[V_1:\mathbf{u}/V_1 \cap V_2:\mathbf{v}, V_2:\mathbf{v}/V_1 \cap V_2:\mathbf{v}])$

The first six equations of the algorithm are the same as in [MM82], except for additional side conditions which make sure that $t$ is not a labeled variable. The interesting cases are 8 and 9. In 8 a pronoun is bound to an antecedent and in 9 two pronouns are identified, i.e., they have the same possible antecedents, namely those which are accessible for both of them. This is accomplished by identifying the pronoun variables and substituting the set of possible antecedents by the intersection of the possible antecedents of each pronoun.

Identification of pronouns underlies different constraints than binding a pronoun to a proper antecedent. To identify two pronouns $\mathbf{u}$ and $\mathbf{v}$, it is not required that $\mathbf{u}$ is accessible from $\mathbf{v}$, or the other way around. But they can only be identified if they have at least one proper accessible antecedent in common.

(16) Buk is a poet. For every man there is a woman who hates him.
   $\models_a$ There is a woman who hates him.
(17) $p(b) \land \forall x(w(x) \to \exists y(w(y) \land ?\mathbf{u}\, h(y, \mathbf{u})))$
   $\models_a \exists z(w(z) \land ?\mathbf{v}\, h(z, \mathbf{v}))$

For instance, in (16) the conclusion is only valid if the first and the second occurrence of *him* are identified. In Section 2 it was said that universal quantification is a barrier for flexible binding, and therefore the second occurrence of *him* cannot be bound to the first one. On the other hand, both of them have a proper antecedent in common, namely the constant $b$ representing the proper name *Buk*. In addition, the first occurrence of *him* has the variable *x* as an accessible antecedent, introduced by the universal quantification *every man*. If one wants to identify them, one has to take the intersection of both sets of accessible antecedents and hence drop *x* as a possible antecedent. Observe that identification of pronouns still leaves some space for underspecification, because the intersection of two pronouns does not have to be a singleton. Of course, identifying two pronouns, where more than one antecedent is accessible for both, forces them to be bound to the same element of the intersection. Both can be bound to any element of the intersection, but it has to be the same one for both pronouns.

If the unification algorithm terminates successfully for a pair of literals P,Q, the solved set determines a substitution $\sigma$ that is the mgu* of P,Q:

$$\sigma := \{s/t \mid s \approx t \in \mathsf{unify}^*(P,Q)\}.$$

A set of equations $\{s_1 \approx t_1, \ldots, s_n \approx t_n\}$ is called *solved* if

1. $s_i \in VAR \cup LPVAR$ and the $s_i$ are pairwise disjoint
2. no $s_i$ occurs in a term $t_j$ $(1 \leq i, j \leq n)$.

**Lemma 6 (Correctness of the Unification* Algorithm).** *Let E be a set of equations and* $\mathsf{unify}^*(E) = E'$, *then*

*(i) E is unifiable* iff $E'$ is unifiable**
*(ii)* $\sigma$ *is the mgu* of E iff $\sigma$ is the mgu* of $E'$*

*Proof.* (i) We have to show that actions 2, 4, 5, 7, 8, and 9 preserve unifiability*, when unify* is applied to a unifiable* set *E*. For 2, 4, and 5, this is obvious. To show it for 7, note that $\tau := [x/t]$ is a unifier* of $x$ and $t$. If $\sigma$ is a unifier* of $\{x \approx t\} \cup E$ then $\sigma$ is of the form $\tau\rho$. Because $\tau\tau = \tau$, it holds that $\sigma = \tau\rho = \tau\tau\rho = \tau\sigma$. Therefore $\sigma$ unifies* $\{x \approx t\} \cup E$ iff $\sigma$ unifies* $\{x \approx t\} \cup E[x/t]$. 8 is analogous to 7, plus the additional side condition that $x \in V$. The last case is 9. If $\{V_1:\mathbf{u} \approx V_2:\mathbf{v}\} \cup E$ is unifiable*, then it is with a unifier* $\sigma$ of the form $\tau\rho$ with $\tau := [V_1:\mathbf{u}/V_1 \cap V_2:\mathbf{v}, V_2:\mathbf{u}/V_1 \cap V_2:\mathbf{v}]$.
Again, $\sigma = \tau\rho = \tau\tau\rho = \tau\sigma$ and then $\sigma$ also unifies*
$\{V_1:\mathbf{u} \approx V_1 \cap V_2:\mathbf{v}, V_2:\mathbf{v} \approx V_1 \cap V_2:\mathbf{v}\} \cup E[V_1:\mathbf{u}/V_1 \cap V_2:\mathbf{v}, V_2:\mathbf{v}/V_1 \cap V_2:\mathbf{v}]$.

(ii) The actions 2, 4, 5, 7, and 8 turn a set of equations into an equivalent one. For $\sigma$ to be the mgu* of $\{V_1:\mathbf{u} \approx V_2:\mathbf{v}\} \cup E$ means according to our definition that $\sigma$ has to be of the form $\tau\rho$, where

$$\tau := [V_1:\mathbf{u}/V_1 \cap V_2:\mathbf{v}, V_2:\mathbf{u}/V_1 \cap V_2:\mathbf{v}].$$

But then $\sigma$ is also the mgu* of

$$\{V_1:\mathbf{u} \approx V_1 \cap V_2:\mathbf{v}, V_2:\mathbf{v} \approx V_1 \cap V_2:\mathbf{v}\} \cup E[V_1:\mathbf{u}/V_1 \cap V_2:\mathbf{v}, V_2:\mathbf{v}/V_1 \cap V_2:\mathbf{v}]. \square$$

**Lemma 7 (Termination of the Unification* Algorithm).** *The unification* algorithm terminates for each finite set of equations.*

*Proof.* If rules 3 and 6 are applied, we are done. Otherwise, rule 7 can be applied only once, because after application the side condition is no longer fulfilled. In 9 it is presupposed that $V_1 \cap V_2$ is a proper subset of $V_2$; this ensures that an application of 9 really reduces the set of possible antecedents. Because 9 can be applied only a finite number of times, it can reintroduce a term $V:\mathbf{u}$ only finitely often, therefore rule 8 can also be applied only finitely many times. Rules 1, 5, and 6 are only applied once, and the number of possible applications of rule 2 is finite as well, because terms contain only finitely many symbols. Therefore all rules can be applied only finitely many times, and termination follows. □

**Proposition 8 (Total Correctness of the Unification* Algorithm).**
*The unification* algorithm computes for each finite set of equations E a solved set, that has the same mgu* as E in finitely many steps iff E is unifiable*.*

*Proof.* The fact that the unification* algorithm preserves unifiability* and that it terminates has been proven in Lemma 1 and 2, respectively. It remains to be shown that the set of equations computed by the algorithm is a solved set. In 7, 8, and 9, the left side of the equation is always substituted in $E$ by the right side of the equation. If the left side is identical to the right side, the equation is erased by rule 4. Therefore, no left side of an equation occurs somewhere else. □

**The Resolution Method.** Having defined labeled unification, it is straightforward to adapt the resolution principle. The only thing we have to change is to make sure that variable disjointness applies only to proper variables (elements of *VAR*). The function *VAR* returns the set of proper variables, when applied to a set of clauses $\Delta : VAR(\Delta) = \{x \in VAR \mid x \text{ occurs in } \Delta\}$. The resolution rule accomplishing pronoun binding ($res_p$) is defined as follows:

$$\frac{C \cup \{\neg P_1, \ldots, \neg P_n\} \quad D \cup \{Q_1, \ldots, Q_m\}}{(C \cup D\pi)\sigma} \ (res_p)$$

where
- $Q_1, \ldots, Q_m$ are atomic
- $\pi$ is a substitution such that
  $VAR(C \cup \{\neg P_1, \ldots, \neg P_n\}) \cap (VAR(D \cup \{Q_1, \ldots, Q_m\}))\pi = \emptyset$
- $\sigma$ is the mgu* of $\{P_1, \ldots, P_n, Q_1\pi, \ldots, Q_m\pi\}$

**Definition 9 (The Proof Algorithm).** Our proof algorithm prf consists of three steps:

1. annotate the conjunction of the premises and the negation of the conclusion;
2. apply clause form transformation; and
3. apply the resolution rule until a contradiction can be derived, or no new resolvents can be generated.

**An Example.** We will only give a very short, and therefore very simple example of a labeled resolution derivation. We hope that it illustrates some of the aspects of labeled resolution mentioned before.

Consider example (16) again, here repeated as (18), where (19) is the corresponding semantic representation.

(18) Buk is a poet. For every man there is a woman who hates him.
    $\models_a$ There is a woman who hates him.

(19) $p(b) \land \forall x(w(x) \rightarrow \exists y(w(y) \land ?\mathbf{u}\, h(y,\mathbf{u})))$
  $\models_a \exists z(w(z) \land ?\mathbf{v}\, h(z,\mathbf{v}))$

**Annotating (19):**
  $\mathsf{annot}(\emptyset, p(b) \land \forall x(w(x) \rightarrow \exists y(w(y) \land ?\mathbf{u}\, h(y,\mathbf{u}))) \land \neg \exists z(w(z) \land ?\mathbf{v}\, h(z,\mathbf{v}))) =$
  $p(b) \land \forall x(w(x) \rightarrow \exists y(w(y) \land h(y, \{b,x\}:\mathbf{u}))) \land \neg \exists z(w(z) \land h(z, \{b\}:\mathbf{v})))$

**Clause form transformation:**

$\{p(b^b)\}, \{m(h^h)\}, \{\neg m(x^x), w(f^y)\}, \{\neg m(x^x), h(f^y, \{b,x\}:\mathbf{u})\}, \{\neg w(z^z), \neg h(z^z, \{b\}:\mathbf{v})\}$, where the additional clause $\{m(h^h)\}$ stems from the assumption that the domain of men is nonempty.

**Resolution:**

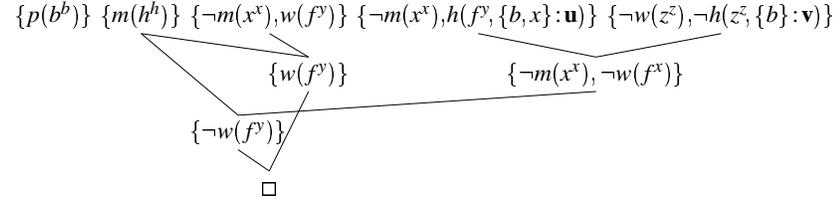

Actually, the only remarkable step in the derivation is resolving
  $\{\neg m(x), h(f, \{b,x\}:\mathbf{u})\}$ and $\{\neg w(z), \neg h(z, \{b\}:\mathbf{v})\}$
with $\{\neg m(x), \neg w(f)\}$ as the resolvent. Here, the two labeled pronoun variables can be identified, because the intersection of their accessible antecedents is nonempty. The corresponding mgu* of
  $\{\neg m(x^x), h(f^y, \{b,x\}:\mathbf{u}), \neg w(z^z), \neg h(z^z, \{b\}:\mathbf{v})\}$
is $\sigma := [x^x/z^z, z^z/f^y, \{b,x\}:\mathbf{u}/\{b\}:\mathbf{v}]$.

Note also, that although $p(b)$ introduced the antecedent $b$, it is not used in the derivation because all information that is necessary to derive the contradiction is captured by the labels. This is the advantage of using labels; it allows us to express non-local dependency relations in our framework, which is essential for dealing with pronoun binding in dynamic semantics where a pronoun and its antecedent can occur in different formulas.

**Evaluation from a Linguistic Point of View.** In general, it is not enough if one gives just the information that there is a binding that allows to derive a conclusion, but one also wants to know *which* binding. It is easy to augment our method in a way such that it accomplishes this simply by memorizing the substitutions of pronoun variables that occur during a derivation.

From a linguistic point of view, one is also interested in comparing different bindings. If we force the proof procedure to backtrack every time it has found a binding which allows to derive a contradiction, we can generate all possible bindings. Probably some of the bindings are preferable to others by taking linguistic heuristics for pronoun resolution into account, see for instance [GJW95], but this is beyond the scope of the present paper.

### 3.3 Results

Before we prove completeness and soundness of our method, we have to explain what these notions mean in our setting.

To show that the resolution principle is correct we have to find the right loop invariant. We will show that if the parent clauses of a resolution step are strongly satisfiable, then so is the resolvent.

**Definition 10 (Strong Satisfiability).** We say that a clause $C$ is *strongly satisfiable* if there is a model $M$ and for all substitutions $\theta$ from *PVAR* to *VAR*), then there is a literal $L \in C\theta$, such that $M \models L$.

**Lemma 11.** *Let $C \cup \{\neg P_1, \ldots, \neg P_n\}$ and $D \cup \{Q_1, \ldots, Q_m\}$ be variable disjoint and strongly satisfiable. If $\sigma$ is the mgu* of $\{P_1, \ldots, P_n, Q_1, \ldots, Q_m\}$, then $C\sigma \setminus \{\neg P_1\sigma\} \cup D\sigma \setminus \{P_1\sigma\}$ is strongly satisfiable.*

*Proof.* The set of possible disambiguations of the resolvent is a subset of the possible disambiguations of the parent clauses, because possible antecedents are unified, and in case of pronoun unification only the intersection of possible antecedents has to be considered. Now, two cases have to be distinguished.

(i) $M \not\models P_i\sigma$. Because $P_i\sigma$ is an instance of $P_i$ and $D \cup \{Q_1, \ldots, Q_m\}$ is strongly satisfiable, it holds that $M \models D\sigma \setminus \{P_1\sigma\}$. But $D\sigma \setminus \{P_1\sigma\}$ is a subset of the resolvent and therefore $M \models C\sigma \setminus \{\neg P_1\sigma\} \cup D\sigma \setminus \{P_1\sigma\}$.

(ii) $M \models P_i\sigma$. Again, $P_i\sigma$ is an instance of $P_i$ and $C \cup \{\neg P_1, \ldots, \neg P_n\}$ is strongly satisfiable. Hence, it holds that $M \models C\sigma \setminus \{\neg P_1\sigma\}$ and thereby $M \models C\sigma \setminus \{\neg P_1\sigma\} \cup D\sigma \setminus \{P_1\sigma\}$. □

**Corollary 12 (Soundness).** *If `prf` (see Definition 9) produces the empty clause on input $\neg \varphi$, then $\varphi$ is valid.*

*Proof.* If we can derive $\square$ from a set of clauses $C$, where C is the clause form of $\neg \varphi$, then we can show by induction that $C$ is not strongly satisfiable, i.e., there is no model $M$ such that $M \models C\theta$ for all possible substitutions. Hence, for all models $M$, there is a disambiguation $\delta$, such that $M \not\models \delta(\neg \varphi)$, which is equivalent to $M \models_a \varphi$, the definition of $\varphi$ being valid. □

**Lemma 13.** *Let $\theta$ be a total disambiguation of $\varphi$, and assume that $\varphi\theta$ is unsatisfiable. Then there is a (classical) resolution deduction of $\square$ from $\varphi\theta$.*

**Lemma 14.** *Let $\theta$ be a total disambiguation of $\varphi$. If there is a resolution deduction of $\square$ from $\varphi\theta$, then `prf` generates the empty clause on input $\varphi$.*

*Proof.* The idea of the proof is to turn the classical resolution proof of $\square$ from $\delta(\varphi)$ into a labeled resolution proof of $\square$ from the original formula $\varphi$ by repeating the resolution steps and inserting the required substitutions (i.e., partial disambiguations) just before any steps where they were used in the original proof.

Although the idea of this proof is simple, the details are too numerous to be included here. □

**Corollary 15 (Completeness).** *If $\varphi$ is valid, then `prf` generates the empty clause on input $\neg \varphi$.*

## 4  Related Work

Most work in the area of ambiguity and discourse semantics focuses on representational issues, but see [vEJ96,MdR98] for calculi for quantificational ambiguities. Approaches that deal with pronoun binding are mostly trying to bind pronouns by applying some heuristics. The work that is closest to ours is the approach of Kohlhase and Konrad [KK98] who deal with pronoun binding in the setting of natural language corrections by using higher-order unification, and a higher-order tableaux method [Koh95] to reason about possible bindings. Van Eijck [vE98] presents a sequent calculus for *DPL* which deals with some of the complications we avoided in this paper; for instance multiple quantification of the same variable. Some of the ways in which dynamic updating can restrict possible pronoun bindings are considered in [Mon98].

## 5  Conclusion

In this paper we have presented a resolution calculus for reasoning with ambiguities triggered by pronouns and the different ways to bind them. Deduction steps and pronoun bindings are interleaved with the effect that only pronouns that are used during a derivation are bound to a possible antecedent. Labels allow us to capture relevant structural information of the original formula on a very local level, namely by annotating variables. Therefore structural manipulation, a prerequisite of any efficient proof method, does no harm.

Our ongoing work focuses on two aspects. First, we have to see how our resolution method behaves when other strategies restricting the search space are added; e.g., set-of-support strategy, ordered unification, or subsumption checking. Second, we are in the process of implementing the annotation and unification$^*$ algorithms and are trying to integrate them into a resolution theorem prover.

**Acknowledgment.**  The research in this paper was supported by the Spinoza project 'Logic in Action' at ILLC, University of Amsterdam.